\documentclass[runningheads]{llncs}
\usepackage{graphicx}
\usepackage{amsmath,amssymb} 
\usepackage{color}

\makeatletter
\newcommand{\printfnsymbol}[1]{%
  \textsuperscript{\@fnsymbol{#1}}%
}
\makeatother 

\usepackage{booktabs}
\usepackage{cite} 
\usepackage{hyperref}

\begin{document}
\pagestyle{headings}
\mainmatter

\def\ACCV20SubNumber{31}  

\title{Speech2Video Synthesis with 3D Skeleton Regularization and Expressive Body Poses} 
\titlerunning{Speech2Video}

%
\author{Miao Liao\thanks{Equal contribution}\and
Sibo Zhang\printfnsymbol{1} \and
Peng Wang \and
Hao Zhu \and
Xinxin Zuo \and
Ruigang Yang}
\authorrunning{M. Liao et al.}
%

\institute{Baidu Research, Baidu Inc \\
\email{\{miao.liao, sibozhang1, jerryking234\}@gmail.com, zhuhaoese@nju.edu.cn, 
\{xinxin.zuo, ryang2\}@uky.edu
}
}

\maketitle
\begin{abstract}
In this paper, we propose a novel approach to convert given speech audio to a photo-realistic speaking video of a specific person, where the output video has synchronized, realistic, and expressive rich body dynamics.
We achieve this by first generating 3D skeleton movements from the audio sequence using a recurrent neural network (RNN), and then synthesizing the output video via a conditional generative adversarial network (GAN).
To make the skeleton movement realistic and expressive, we embed the knowledge of an articulated 3D human skeleton and a learned dictionary of personal speech iconic gestures into the generation process in both learning and testing pipelines. 
The former prevents the generation of unreasonable body distortion, while the later helps our model quickly learn meaningful body movement through a few recorded videos. 
To produce photo-realistic and high-resolution video with motion details, we propose to insert part attention mechanisms in the conditional GAN, where each detailed part, e.g. head and hand, is automatically zoomed in to have their own discriminators.
To validate our approach, we collect a dataset with 20 high-quality videos from 1 male and 1 female model reading various documents under different topics. Compared with previous SoTA pipelines handling similar tasks, our approach achieves better results by a user study. 
\keywords{Face, Gesture, and Body Pose, Image and Video Synthesis, Vision and Language, GAN}
\end{abstract}

\section{Introduction}

\textit{Speech2Video} is a task of synthesizing a video of human full body movements, including head, mouth, arms etc., from a speech audio input. The produced video should be visually natural and consistent with the given speech. 
Traditional way of \textit{Speech2Video} involves performance capture with dedicated devices and professional operators, and most of the speech and rendering tasks are performed by a team of animators, which is often costly for custom usage. 
Recently, with the successful application of deep neural networks, data-driven approaches have been proposed for low cost speech video synthesis. For instances, SythesisObama~\cite{suwajanakorn2017synthesizing} or MouthEditting~\cite{fried2019text} focus on synthesizing a talking mouth by driving mouth motion with speech using an RNN.
Taylor \cite{taylor2017deep} propose to drive a high fidelity graphics model using audio, where not only animates mouth but also other parts on the face are animated to obtain richer speech expressions. 

However, mouth movement synthesis is mostly deterministic, given a pronunciation, the movement or shape of the mouth is similar across different persons and context. In our task of \textit{Speech2Video}, a full body gesture movement under the same situation is more generative and has more variations, the gestures are highly dependent on current context and individual person who is doing the speech. Personalized speaking gestures appear at certain moment when delivering important messages. Therefore, useful information is only sparsely present in a video, yielding difficulties for a simple end-to-end learning algorithm~\cite{suwajanakorn2017synthesizing,taylor2017deep} to capture this diversity from the limited recorded videos. 

To the best of our knowledge, LumiereNet~\cite{kim2019lumi} is the most related work performing a similar task with ours, which builds an end-to-end network for full upper body synthesis. However, in their experiments, the body motion is less expressive where the major dynamics are still at the talking head. 
In practice, following a similar method, we build a pipeline for body synthesis, and train it with our collected online speech videos, where three major issue exists. First, as discussed, the generated body movements only have repetitive patterns, while the ground truth video contains emphasis gestures at certain moments. Second, the generated body appearance at detailed parts such as hand and elbow could be unnaturally distorted, which is geometrically infeasible.
Last, the generated body and hand appearance are blurry with motions.
Therefore, in this work, we propose a novel trainable \textit{Speech2Video} pipeline, which handles all these issues simultaneously. For handling diversity issues, we build a pose dictionary with text for each person from their presentation videos. To guarantee the generated pose are physical plausible, we enforce the 3D skeleton as the intermediate representations, the generated joints should follow the regularity of 
human body. Finally, to ensure high quality synthesized appearance, we propose an part-aware discriminator to provide additional attention of generated detailed parts like arms and hands. 



Finally, in order to better evaluate our system, we create a dataset with recorded speech videos of several target  while they are reading some carefully selected articles, using camera with high resolution and frame rate (FPS). 
In our experiments, we show our approach generates perceptually  better human dynamics than other existing pipelines with more gesture variations. 
%

The main contributions of this paper are summarized as follows:
\begin{itemize}
\item We proposed a novel 2-stage pipeline of generating an audio-driven virtual speaker with full-body motions including the face, hand, mouth and body. Our 3D driven approach overcomes issues of direct audio-to-video approach where human appearance details are missing. And it also makes it possible to insert key poses in the human motion sequence. It is shown in the result section why we have to decompose this task into a 2-stage generation, instead of direct audio-to-video generation. 
\item A dictionary of personal key poses is built that adds more dimensions to the generated human poses. Besides, we presented an approach to insert key poses into the existing sequence.
\item 3D skeleton constraints are embedded to generate body dynamics, which guarantees the pose is physically plausible.
\item We proposed a modified GAN to emphasize on face and hands to recover more details in the final output video.
\end{itemize}


\section{Related Work}

\textbf{Human Body Pose Estimation and Fitting}  \cite{ge20193d} proposed 3d shape and pose estimation specific for hands. \cite{kanazawa2019learning, pavllo20193d} predicts 3d human motion from video or a single image, but they are limited to fit human model with limb only, not hands or face. While openpose~\cite{cao2018openpose} has been so successful at fitting the detailed human model to 2D image with all our demanded parts including face and fingers, their output is 2D landmarks in the image space. Based on openpose, SMPL-X~\cite{SMPL-X:2019} fits a 3D skeleton to those output 2D landmarks through an optimization. It also parameterizes human motion as joint angles, making it much easier to constrain joints under reasonable human articulation.  

\noindent\textbf{Audio to Motion} \cite{karras2017audio} drives high fidelity 3D facial model using audio via end-to-end learning, where both poses and emotions are learned. \cite{shlizerman2018audio} focuses on synthesizing hand motion from music input, rather than speech. Its goal is to animate graphics models of hands and arms with piano or violin music. \cite{yan2019convolutional} generates skeleton-based action using Convolutional Sequence Generation Network (CSGN). \cite{martinez2017human} instead, predict human motion using recurrent neural networks. \cite{li2017auto} uses auto-conditioned recurrent networks for extended complex human motion synthesis. They can model more complex motions, including dances or martial arts. 

\noindent\textbf{Video Generation from Skeleton} 
pix2pix~\cite{isola2017image, wang2018pix2pixHD} is a milestone in the development of GAN. It outputs an detailed real-life image from an input semantic label image. In our pipeline, the semantic label maps are image frames of the human skeleton. Nevertheless, direct applying pix2pix to an input video without temporal constraints will result in incoherent output videos. Therefore, vid2vid~\cite{wang2018vid2vid} is proposed to enforce temporal coherence between neighboring frames. 
\cite{shysheya2019textured} proposes to render realistic video from skeleton models without building a 3D model, where the second stage of video generation is emphasized. However, it doesn't take care of facial expression and mouth movement, and it doesn't address the problem of how to generate realistic movement of the skeleton body model. \cite{cai2018deep} proposes a similar pipeline, which generates skeleton pose first and then generate the final video. However, rather than audio, its input is random noise and its skeleton model is very simple only having body limbs. That means its final output video lacks details on the face and fingers. 

\noindent\textbf{Character Synthesis} \cite{suwajanakorn2017synthesizing, fried2019text, mittal2020animating} focus on synthesizing a talking head by driving 2D mouth motion with speech. When the mouth sequence is generated via texture mapping, it is pasted onto an existing video after lighting and texture fusion. \cite{thies2019neural} instead, drives a 3D face model by audio, and render the final video using a technique called neural renderer~\cite{thies2019deferred}. \cite{kim2019lumi} attempts to produce videos of an upper-body of a virtual lecturer, but the only moving part is still the mouth. Face2Face~\cite{thies2016face2face} transfers expressions from a person to a target subject using a monocular RGB camera. Given a video of a dancing person, \cite{chan2019everybody} transfers the dancing motion to another person, even though the second person does not know how to dance. The second person is only required to record a video of a few poses. While achieving good results, there are still visible distortion and blurriness on the arms, not to mention details of hands. Liquid Warping GAN~\cite{liu2019liquid} is a recent work to synthesize human videos of novel poses, viewpoints, and even clothes. They have achieved decent results given that their input is simply a single image. Their work is mainly focused on image/video generation, while our main contribution is simulating human motions. \cite{cai2018deep} proposed a similar pipeline as ours, which generate skeleton pose first and then generate the final video. However, rather than audio, its input is random noise and its skeleton model is very simple only having body limbs. That means its final output video lacks details on the face and fingers. In contrast, our skeleton model consists of limbs, face, and fingers. \cite{ginosar2019learning} learns individual styles of speech gesture via 2 stages as we propose, but its rendering part produces quite a few artifacts in the final generated videos.

\section{Methods}

\begin{figure*}[t]
  \centering
  \includegraphics[width=4in]{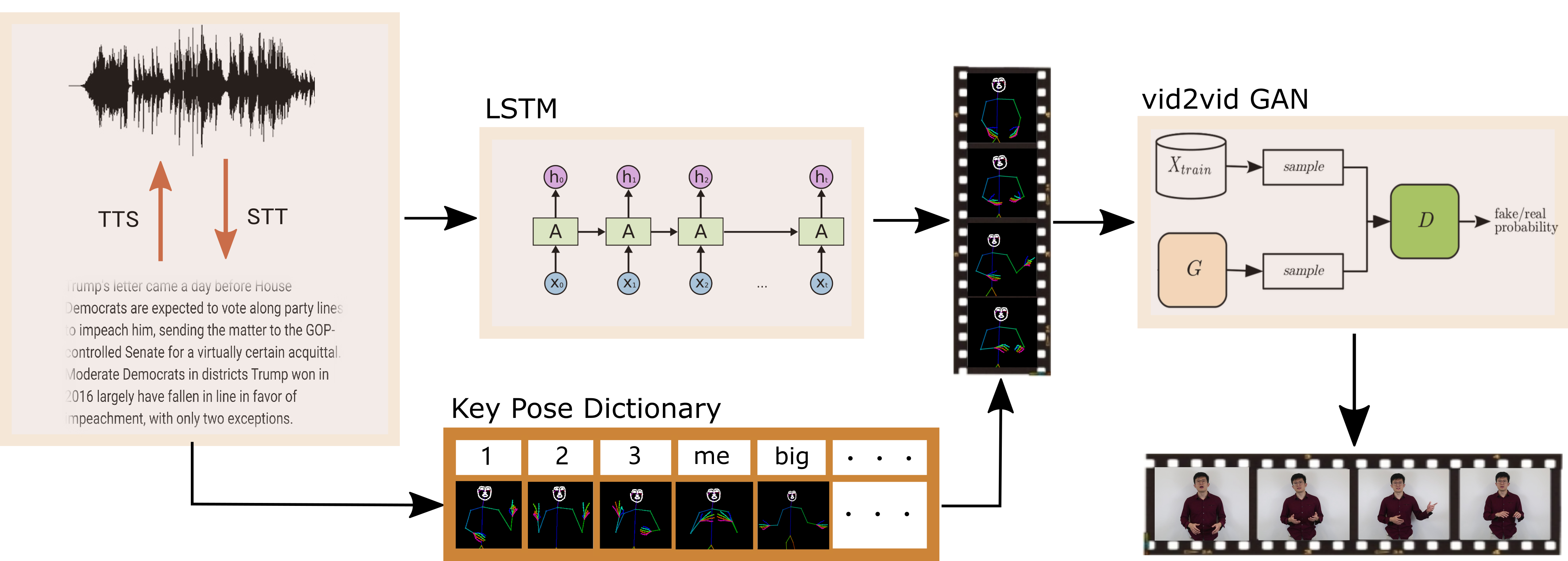}
  \caption{Pipeline of our system.}
  \label{fig:pipeline}
\end{figure*}

As shown in figure~\ref{fig:pipeline}, the input to our system is audio or text, depending on what is used to train the long short-term memory (LSTM) network. We here assume that audio and text are interchangeable, given both text-to-speech (TTS) and speech-to-text (STT) technologies are mature and commercially available. Even though we still get some wrongly recognized words/characters from the state of the art STT engine, our system can tolerate these errors quite successfully,  because the main purpose of this LSTM network is to map texts/audios to body shapes. Wrong STT outputs are usually words with similar pronunciations to those of the true ones, meaning they are very likely to have similar spelling too. Therefore, they will eventually map to more or less alike body shapes.

The output of the LSTM is a sequence of human poses, parametrized by SMPL-X~\cite{SMPL-X:2019}. SMPL-X is a joint 3D model of the human body, face, and hands together. This dynamic joint 3D model is visualized as a sequence of 2D colorized skeleton images. These 2D images are further input into a vid2vid generative network~\cite{wang2018vid2vid} to generate final realistic people images. 

We found that while successfully synchronize speech and movement, LSTM learns only repetitive human motions most of the time, which results in boring looking videos. In order to make the human motion more expressive and various, we insert certain poses into the output motions of LSTM when some key words are spoken, for example, huge, tiny, high, low, and so on. We manually build a dictionary that maps those key words to their corresponding poses. Please refer to the following sections for details on how we build this dictionary. 

\begin{figure*}[t]
  \centering
  \includegraphics[width=4in]{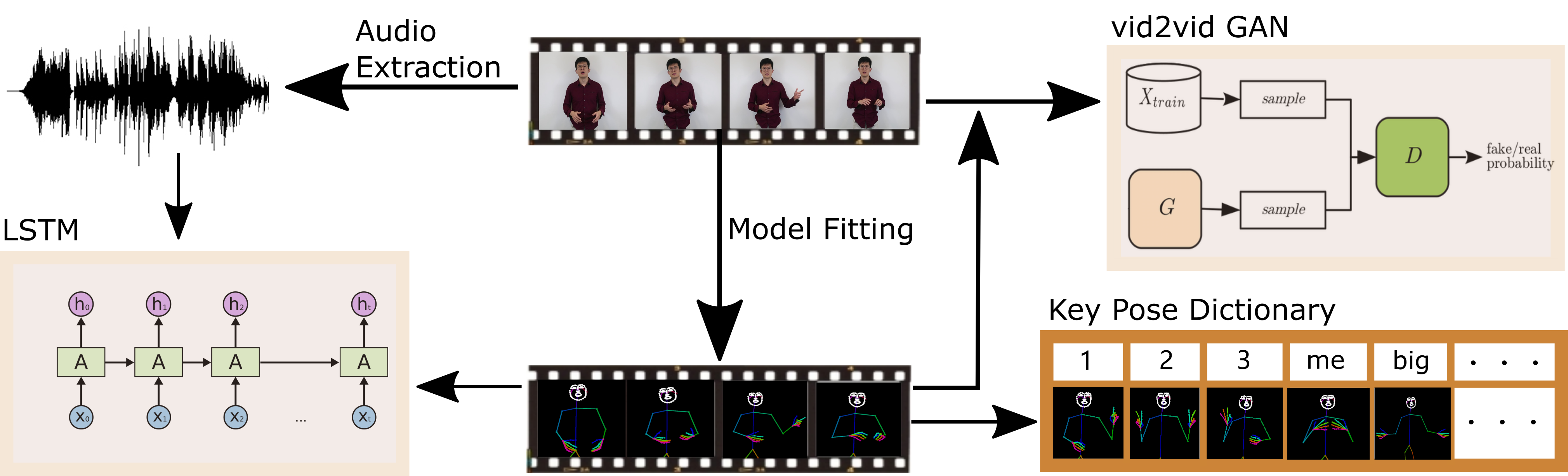}
  \caption{Overview of our training process.}
  \label{fig:training}
\end{figure*}

Training the LSTM and vid2vid networks requires only some videos of target animation subject reading a script. As shown in figure~\ref{fig:training}, given a video of a talking person, we first fit a human body model to each frame. Together with the extracted audio on the left-hand side, it is fed into the LSTM to train mapping from audio to human poses. On the right-hand side, 2D skeleton images of the human body model and their corresponding true person images are used to train vid2vid generative network. Finally, we manually select some key poses and build a dictionary that maps key words to key poses.

\subsection{Speech2Video dataset}

Ideally, our system is capable of synthesizing anyone as long as we can download some of their speech videos from websites such as Youtube. In reality, however, most of those Youtube videos are shot under auto exposure mode, meaning the exposure time could be as long as 33 milliseconds for 30 fps videos. It is impossible to capture clear hand images under such long exposure time when the hands are moving. In fact, most of these frames have motion blur to some extent, which causes big problems when we fit the hand finger model to the images. In addition, our system requires our speaker to be present in a constant viewpoint, but a lot of speech videos keep changing their viewpoint. Though fitting to blurry image itself is a good research topic, we only focus on the video synthesis part and use the existing state-of-the-art approach to fit human model. Therefore, we decided to capture our own data. 

We invited two models to capture our training data, one English speaking female and one Chinese speaking male. We capture a total of 3 hours of videos for each model when they were reading a variety of scripts, including politics, economy, sports and so on. We set up our own recording studio with a DSLR camera, which captures 1280 $\times$ 720 videos at 60 frames per second. We fix the exposure time at 5 milliseconds so that no motion blur will be present in the frames. In order to reduce data size, we sample every 5 frames from the video and only work on this subset data. 

\begin{figure}[t]
  \centering
  \includegraphics[width=0.9\linewidth]{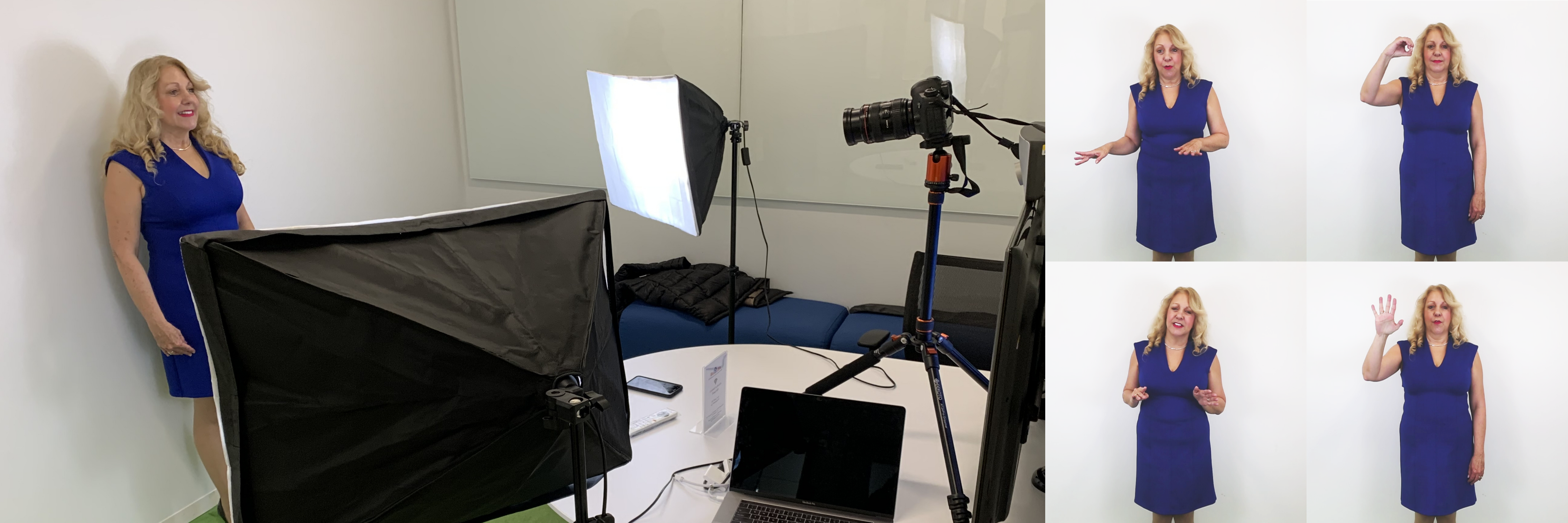}
  \caption{Left: our data capture room. Right: 4 frames from captured video.}
  \label{fig:capture}
\end{figure}

Figure~\ref{fig:capture} shows our data capture room. Our model stands in front of a camera and screen, and we capture a few videos while he/she reads scripts on the screen. In the end, we ask our model to pose for certain key words, such as huge, tiny, up, down, me, you, and so on.

\subsection{Body Model Fitting}
Fitting a human body model to images is equivalent to detecting human keypoints. OpenPose~\cite{cao2018openpose} has done an excellent work on this.  It is a real-time approach to detect the 2D pose of multiple people in an image, including body, foot, hand, and facial keypoints.

\begin{figure}[t]
  \centering
  \includegraphics[width=0.7\linewidth]{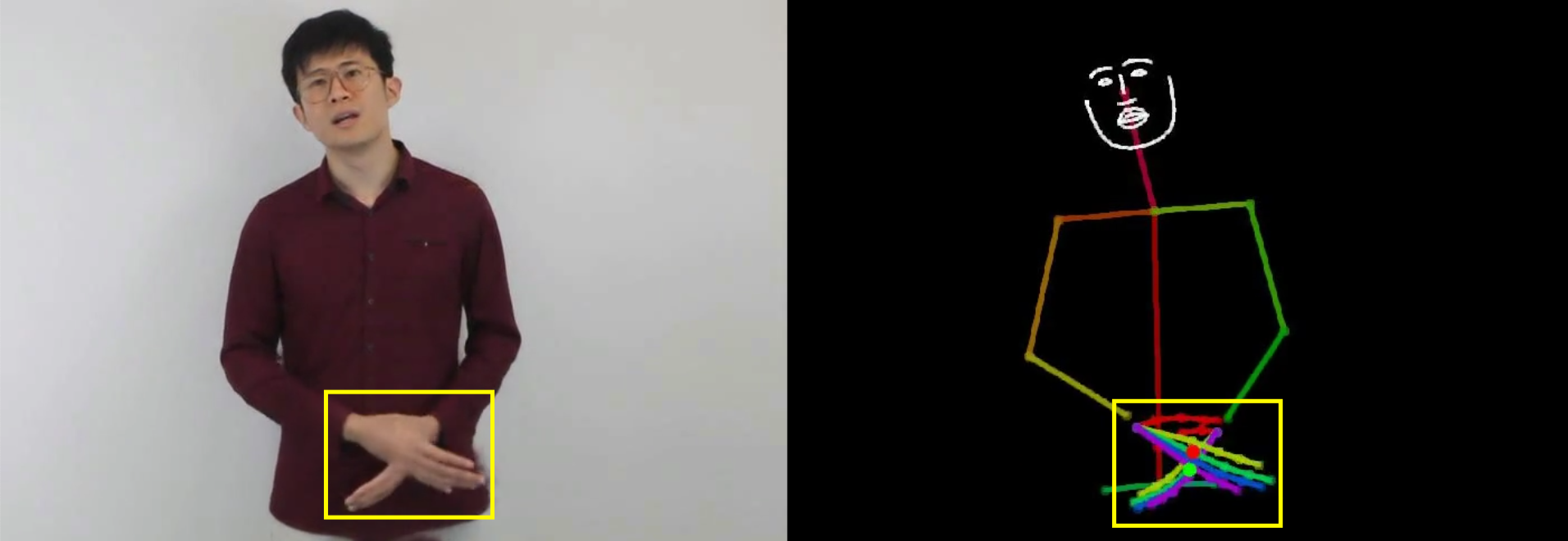}
  \caption{Failure case with 2d model: elongated fingers. }
  \label{fig:failure}
\end{figure}

We first attempted to take those 2D keypoints as a representation of our human body model, and trained the LSTM network that generates 2D positions of these keypoints from audio inputs. The results were not quite satisfactory due to the distortion of output arm and hand (shown in figure ~\ref{fig:failure}). This because in this simply 2D keypoint human model, there is no relationship between 2 connected keypoints. They can virtually move to anywhere independently without constraints from other keypoint, leading to elongated or shorter arms and fingers. Furthermore, at the stage of inserting key poses into existing body motion, it involves interpolating between 2 poses. Direct interpolation on 2D keypoints usually results in invalid intermediate poses that violate human articulated structure.

Under these observations, we adopt SMPL-X, a true articulated 3D human model. SMPL-X models human body dynamics using a kinematic skeleton model. It has 54 joints including neck, fingers, arms, legs, and feet. It is parameterized by a function $M(\theta, \beta, \psi)$, where $\theta \in R^{3(K+1)}$ is the pose parameter and K is the number of body joints plus an additional global body orientation. $\beta \in R^{|\beta|}$ is the shape parameter which controls the length of each skeleton bone. Finally, the face expression parameter is denoted by  $\psi \in R^{|\psi|}$. There are a total of 119 parameters in SMPL-X model, 75 of which come from the global orientation as well as 24 joints excluding hands, each denoted by a 3 DoF axis-angle rotation. The joints on hands are encoded separately by 24 parameters in a lower dimensional PCA space, following approach described in MANO~\cite{romero2017embodied}. The shape and face expression both have 10 parameters respectively.

To fit SMPL-X human model to images, in general, we need to find optimal parameters that minimize $E(\theta, \beta, \psi)$, the weighted distance between 2D projection of those 3D joints and 2D detections of the corresponding joints by OpenPose library~\cite{cao2018openpose}. The weights are determined by detection confidence scores, so that noisy detection will have less influence on the gradient direction. In our specific scenario, we modified the fitting code to fix body shape parameters $\beta$ and global orientation during the optimization. Because we are dealing with the same person within a video and the person is standing still during the entire video. We only compute the human body parameter $\beta$ and human global orientation for the first frame and use them for the remaining frames. So the final objective function for us becomes $E(\theta, \psi)$, where we only look for optimal pose and facial expression parameters. That reduces the total number of parameters to 106.

\subsection{Dictionary Building and Key Pose Insertion}

\begin{figure}
  \centering
  \includegraphics[width=0.8\linewidth]{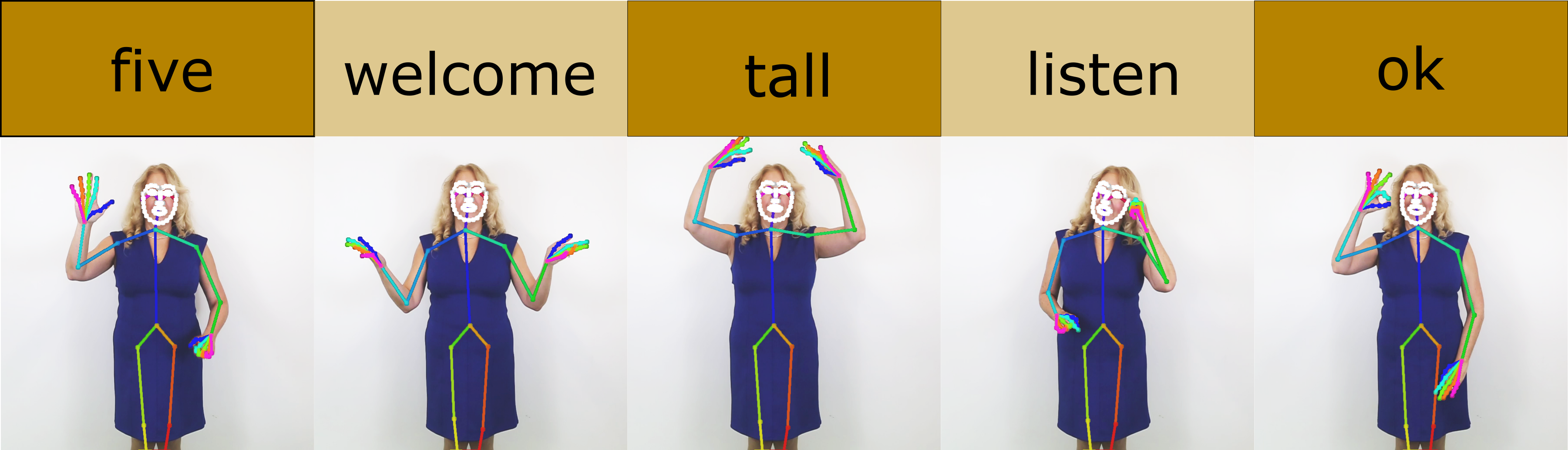}
  \caption{Key words to key poses dictionary. Note a key pose could be a still single frame pose or a multi-frame motion.}
  \label{fig:dict}
\end{figure}

As shown in figure~\ref{fig:dict}, we manually select key poses from the recorded videos and build a word-to-pose lookup dictionary. Again, the pose is represented as 106 SMPL-X parameters. Note that a key pose could be a still single frame pose or a multi-frame motion. We can insert both into an existing human skeleton video by the same approach.

In order to insert a key pose, we first need to know when it's corresponding key word is spoken. For a text-to-speech (TTS) generated audio, the TTS output will include the timestamp of each word in the generated audio. For an audio from a real person, we need to first pass it to a speech-to-text (STT) engine, which generates text script of the speech as well as the timestamp of each individual word. 
We go over all the words within the speech script and look them up in our word-to-pose dictionary. Once they are found in the dictionary, we decide if we want to insert them into the skeleton video by a certain probability. Since some words like "I", "we", "me" could be spoken a few times in a speech. A real person won't pose every time they speak those words. The probability could vary between different words and should be set when we build the dictionary.

\begin{figure*}
  \centering
  \includegraphics[width=0.9\linewidth]{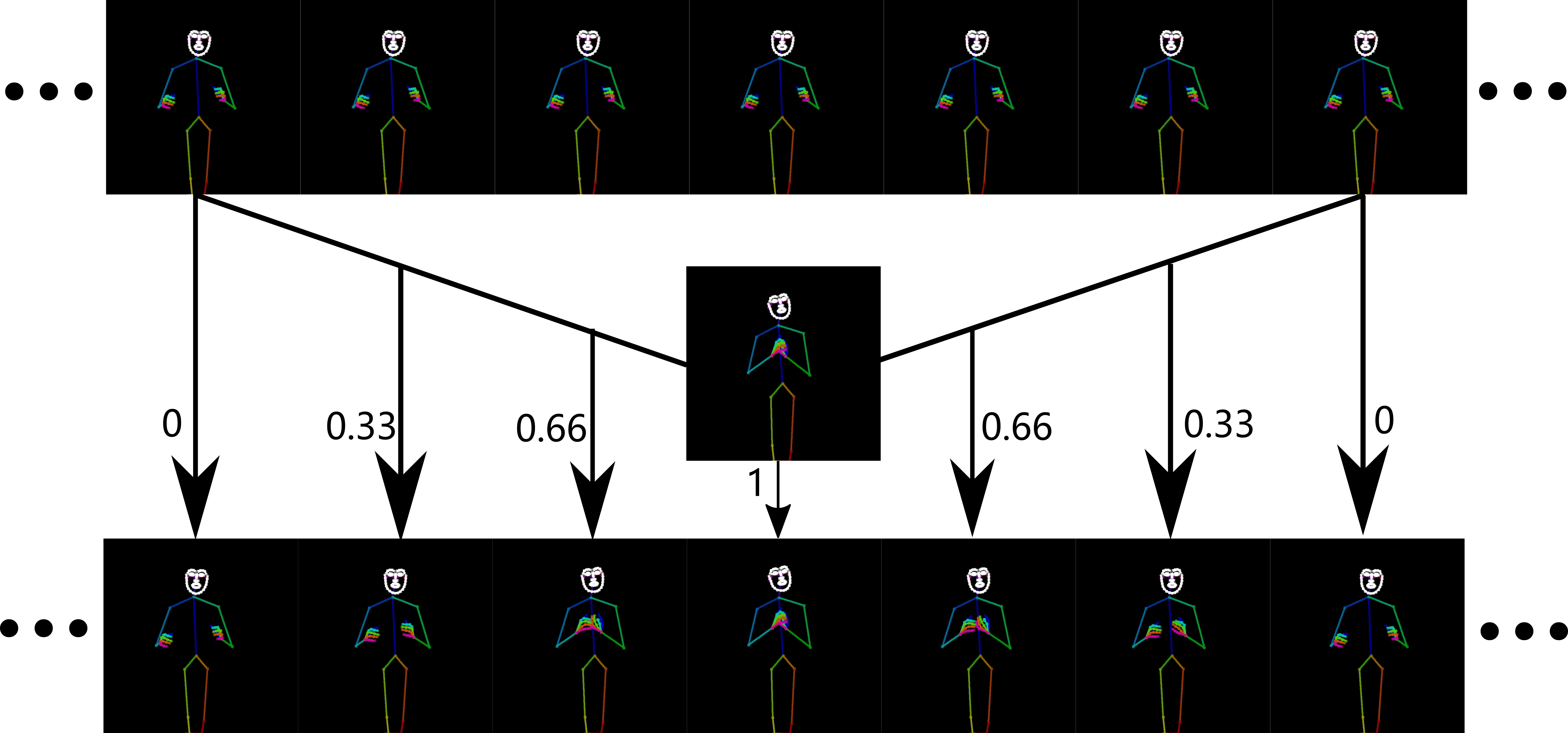}
  \caption{Inserting a key pose smoothly into an existing video sequence. A key pose is inserted into a video with a ramp length N frames before and after its insertion time point. The ramp length is only 3 here for illustration, but the real ramp length is way longer than this. Those number alongside vertical arrows are interpolation weights of the key pose. The weighted sum of ramp start/end pose and key pose replaces original frames in between.}
  \label{fig:insert}
\end{figure*}

When we insert a pose into a video, we do a smooth interpolation in the 106 parameter space. Illustrated in figure~\ref{fig:insert}, a key pose is inserted into a video with a ramp length N frames before and after its insertion time point. The ramp length depends on video frame rate and ramp duration. In all our experiments, the ramp duration is set to be 0.6 seconds. The key pose is directly copied to its time point within the video and overwrite the original frame. In order to maintain a smooth transition to this pose, we also replace frames from ramp start point all the way to the key pose frame on both sides. The new frames are linear interpolated between ramp start frame and key pose frame, weighted by their distance to those 2 frames. 

If our key pose is a single frame still pose, it's inserted exactly as described above, except for one thing. People usually make a pose and keep it for a certain time period. So, instead of showing the key pose in one frame, we also need to keep the key pose for a while. In all our experiments, we keep the pose for 0.3 seconds by duplicating the key pose frame in place multiple times. If our key pose is a motion (a sequence of frames), then it will be copied to the target video to overwrite a sequence of the same length. The smoothness ramping is done the same way.

\subsection{Train LSTM}

When we train the LSTM which maps audio sequence to pose sequence, we have to give different weights to different parts of the human body in the loss, because they have different scales. The relative weights we set to body, hands, mouth, and face are 1, 4, 100, 100 respectively. We also enforce a smoothness constraint on the output pose sequence by adding a difference loss between 2 consecutive poses, in order to make sure the output motion is smooth and natural.

\textbf{Audio to Pose}. We extract the audio features using standard MFCC coefficients~\cite{logan2000mel}. The input audio may have various volume level, we first normalize its volume by RMS-based normalization~\cite{katz2003mastering}. Then for every 25ms-length audio clip, we apply discrete Fourier Transform to get its representation in the frequency domain. The audio clip is sampled at 10ms interval. 40 triangular Mel-scale filters are applied to the output of Fourier Transform, followed by a logarithm operator. Next, we reduce the output dimension to 13 by applying a Discrete Cosine Transform. The final feature is a 28D vector, where the first 14D consists of the 13D output of the Discrete Cosine Transform plus the log mean value of volume, and the second 14D is temporal first-order derivatives of the first 14D, a.k.a, the difference to the previous feature vector.

\textbf{Text to Pose}. Voice could be quite different from people to people, even when they are speaking the same words. That could lead to poor performance of the LSTM learning. Alternatively, we can use text, instead of audio to train the LSTM. That requires us to convert to text if the input is audio. Thanks to the development of natural language processing (NLP), there are quite a few prior works~\cite{reddy1976speech} that do excellent jobs on this. 

For English, we directly use words as the input sequence to LSTM, since word spelling itself incorporates pronunciation information. We pad remaining pausing parts with 0's to form an entire input sequence. On the other hand, for those non-latin languages, for example Chinese, its words/characters don't carry pronunciation information. In this case, we still want to have the same mouth shape and body pose when 2 characters of the same pronunciation are spoken. Therefore, we have to convert characters to representation with phoneme information. For Chinese, we convert each individual character into pinyin, which is composed of 26 English letters. It guarantees 2 characters have the same spelling if they have the same pronunciations.


\textbf{LSTM Architecture}
We opt for a simple 2 layer unidirectional LSTM~\cite{hochreiter1997long}. 
Input is vector of audio/text encoding, and 
 output is vector of SMPL-X parameters. Note that a time delay is applied to the output by shifting output parameter forward in timeline as explored in~\cite{graves2005framewise}. This gives the network the options to predict human poses by looking in the future of speaking. This is especially true when a speaker tends to pose before he/she starting speaking. The dimension of the cell state is set to 300, and the time delay of output is set to 200ms. The network is solved by minimizing a L2-loss on the SMPL-X parameters using Adam optimizer~\cite{kingma2014adam} implemented under TensorFlow. The network is trained with a batch size of 100 and learning rate of 0.001. The input vector is normalized by its mean and variance, but the output is kept unchanged in order to keep the relative scale of different SMPL-X parameters.

\subsection{Train Video Generative Network}
\begin{figure}
  \centering
  \includegraphics[width=0.8\linewidth]{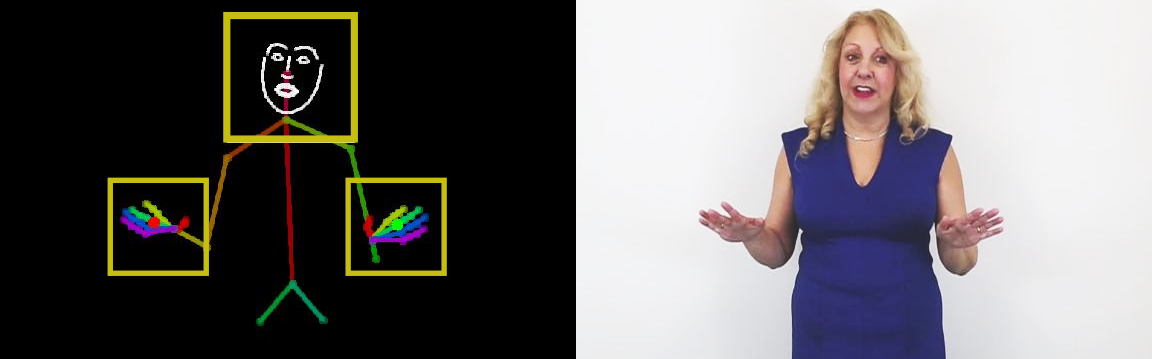}
  \caption{Sample images pair used to train vid2vid. Both hands are labeled by a special color circle. The color circles are identified within the GAN, in order to crop the sub-images around both hands. Those sub-images are passed to discriminator separately from the whole image to ensure we put more weights on the hand detail generation. }
  \label{fig:vid2vid}
\end{figure}

We adopt the generative network proposed by vid2vid~\cite{wang2018vid2vid} to convert our skeleton images into real person images. 
In our applications, the rendering results of human bodies are not equally important. The most important parts are face and hands. To make vid2vid network put more effort on generating details of face and both hands, we modified the network and our input images to achieve this. Specifically, we draw a color circle on both hands on the input skeleton image and also draw face part with white color, which is different from other parts of body (figure~\ref{fig:vid2vid}). Within the network, an image is output from the generative network given an input image. Before we pass it to discriminator network, we locate regions of face and both hands by their special colors in the input image. Then we crop those 3 sub images from the generated image, and pass them to the discriminator network along with the entire output image. The loss weights for those sub images are carefully tuned to make sure the discriminator is more picky on the reality of generated face and hands images.

\textbf{Running times and hardware}.
The most time consuming and memory consuming stage of our system is training the vid2vid network. It takes about a week to finish 20 epochs of training on a cluster of 8 NVIDIA Tesla M40 24G GPUs. The testing stage is much faster. It takes only about 0.5 seconds to generate one frame on a single GPU.

\section{Results}

\subsection{Evaluation and Analysis}
Note it is not straightforward to compare with other methods, because 1) there is no benchmark dataset to evaluate speech to full body videos, 2) people's speech motion is quite subjective and personalized which makes it difficult to define ground truth. As Table~\ref{tab:score} showing, we choose to compare our results with 4 SoTA approaches using user study. We get the best overall quality score compared to other 4 SOTA methods.

\begin{table}[]
\begin{center}
\caption{Average scores of 248 participants on 4 questions. Q1: Completeness of body. Q2: The face is clear. Q3: The body movement is correlated with audio. Q4: Overall quality.}
\label{tab:score}
\begin{tabular}{lllll}
\hline
                      & Q1    & Q2    & Q3    & Q4    \\ \hline
LearningGesture~\cite{ginosar2019learning}       & 3.414 & 3.659 & 3.914 & 3.308 \\
LumiereNet~\cite{kim2019lumi}            & 3.585 & 3.521 & 3.085 & 3.265 \\
Neural-voice-puppetry~\cite{thies2019neural} & 3.202 & 3.840 & 3.180 & 3.542 \\
EverybodyDance~\cite{chan2019everybody}        & 3.944 & 3.662 & 3.680 & 3.681 \\
Our method            & 3.894 & 4.011 & 3.383 & 3.762 \\ \hline
\end{tabular}
\end{center}
\end{table}

\textbf{Inception Score Comparison}.
Also we evaluate our image generation result using inception scores. The score measures two things simultaneously: The image quality and image diversity. 

\begin{table}[]
\begin{center}
\caption{Inception scores for generated videos (IS) and ground truth videos (GT IS) of different methods. The relative incpetion score (Rel. IS) is the ratio of the first to the second. }
\label{tab:inceptionscore}
\begin{tabular}{@{}llllll@{}}
\toprule
     & SynthesizeObama~\cite{suwajanakorn2017synthesizing} \hspace{5mm} & EverybodyDance~\cite{chan2019everybody} \hspace{8mm}& \textbf{Ours}  \\ \midrule
IS & 1.039      &  1.690  & 1.286  \\
GT IS   & 1.127      & 1.818  & 1.351 \\
Rel. IS \hspace{5mm}  & 0.921      & 0.929  & \textbf{0.952}   \\ \bottomrule
\end{tabular}
\end{center}
\end{table}

As shown in Table~\ref{tab:inceptionscore}, we compare to SynthesizeObama and EverybodyDance by computing inception scores on all the frames of videos generated by each method. IS is the score for generated videos and GT IS is the score for ground truth videos. For SynthesizeObama~\cite{suwajanakorn2017synthesizing} the ground truth is the source video of the input audio. For EverybodyDance~\cite{chan2019everybody} the ground truth is the source video to transfer motion from. And in our case, the ground truth is the training video. It is expected that dancing videos (EverybodyDance) have higher scores than speech videos (ours), and speech videos (ours) have higher scores than talking head (SynthesizeObama), since dancing has the most motion varieties. 
Therefore, we use the relative inception scores (inception score of generated videos to ground truth videos) to measure similarity to the ground truth. Our method outperforms the other two methods by this standard, meaning our visual quality is closer to ground truth.

\textbf{Numerical Evaluation}. Since people don't pose exactly the same, even if the same person speaks the same sentence twice. So, it is difficult to tell if our generated body motion is good or not, due to lacking of ground truth. The only part that takes the same shape when speaking the same words is mouth. Thus, we use only mouth to evaluate our motion reconstruction accuracy. Specifically, we record a separate video of our models when they speak totally different sentences than in the training dataset. We extract the audio and input into our pipeline. The output 3D joints of mouth are projected onto the image space and compared to those 2D mouth keypoints detected by OpenPose. The errors are measured by average pixel distance.

\begin{table}[]
\begin{minipage}[t]{.48\textwidth}

\centering
\caption{Numerical evaluation on mouth motion reconstruction of our system. Number here is average pixel distance.}
\label{tab:eval}
\begin{tabular}{@{}llllll@{}}
\toprule
     & Orig. & Man1 & Man2 & Man3 & Text  \\ \midrule
0.5h & 1.769      & 1.838  & 1.911  & 1.992  & 2.043 \\
1h   & 1.730      & 1.868  & 1.983  & 2.012  & 2.024 \\
2h   & 1.733      & 1.809  & 1.930  & 2.047  & 1.993 \\ \bottomrule
\end{tabular}

\end{minipage}%
\hspace{0.4cm}%
\begin{minipage}[t]{.48\textwidth}

\centering
\caption{Average scores of 112 participants on 5 questions. Q1: Completeness of body. Q2: The face is clear. Q3: The human motion looks natural. Q4: The body movement is correlated with audio. Q5: Overall quality.}
\label{tab:video_score}
\begin{tabular}{@{}llllll@{}}
\toprule
             & Q1   & Q2   & Q3   & Q4   & Q5   \\ \midrule
Synth.         & 4.12 & 4.21  & 2.86 & 3.07 & 3.42 \\
TTS   & 4.07 & 3.81 & 2.67 & 2.88 & 3.28 \\
Real & 4.28 & 4.38 & 4.45 & 4.35 & 4.38 \\ \bottomrule
\end{tabular}

\end{minipage}
\end{table}

As reported in table~\ref{tab:eval}, we have done several evaluations on the mouth motion reconstruction and found some interesting facts. We first tried to train our LSTM network using different dataset size to see how it affects the reconstruction accuracy. We used dataset of varying length including 0.5 hour, 1 hour and 2 hours. We use the voice of the same lady (Orig.)  as in training data to do the evaluation. In addition, we also lower the pitch of the original voice to simulate a man's voice, in order to see how voice variation affect the results. We simulate voices of a young man (Man1), a middle age man (Man2) and an old man (Man3) by successively lower pitch values of the original audio. Finally, we train and test our LSTM network using text and compare the results to those of audio.

We have 3 observations from table~\ref{tab:eval}. First, audio has better accuracy than text. Second, longer training dataset doesn't necessarily increase the accuracy for audio but it indeed helps for text. Third, accuracy gets worse when voice is getting more deviated from the original one. The third one is easy to understand, so we expect worse performance if the test voice sounds different from the training voice. For the first and second observations, the explanation is that audio space is smaller than text space, because some words/characters share the same pronunciation, for example, pair vs pear, see vs sea. Therefore, audio training data cover larger parts in its own space than text training data of the same length. In our experiments here, it looks like 0.5-hour length audio is enough to cover the entire pronunciation space. Adding more training data doesn't help increase accuracy. On the other hand, 2-hour length text is still not enough to cover the entire spelling space, so the error keeps decreasing as we increase the length of training data.

\textbf{User Study} To evaluate the final output videos, We conducted a human subjective test on Amazon Mechanical Turk (AMT) with 112 participants. We show a total of five videos to the participants. Four of them are our synthesized videos, two of which are generated by real person audios and the other two are generated by TTS audios. The remaining one is a short clip of a real person. Those five videos are ordered randomly and we didn't tell our participants that there is a real video. The participants are required to rate the quality of those videos on a Likert scale from 1 (strongly disagree) to 5 (strongly agree). Those include  1) Completeness of human body (no missing body parts or hand fingers); 2) The face in the video is clear; 3) The human motion (arm, hand, body gesture) in the video looks natural and smooth; 4) The body movement and gesture is correlated with audio; 5) Overall visual quality of the video and it looks real. 

As shown in table~\ref{tab:video_score}, our synthesized videos (Synth.) get 3.42 and real video gets 4.38 (out of 5). In particular, our proposed method has the same performance on body completeness, face clarity compared to real video. Another discovery is that TTS generated videos are worse than real-audio generated videos in all aspects. The reason is twofold. First, TTS audios are generally more distant to real audios in MFCC feature space, leading to worse reconstructed motions and gestures (conclusion from table~\ref{tab:eval}). Secondly, TTS audio itself sounds fake, which decreases the overall video quality.

\subsection{Ablation Study}

\begin{figure}
\centering
\begin{minipage}[t]{.45\textwidth}
  \centering
  \includegraphics[width=0.8\linewidth]{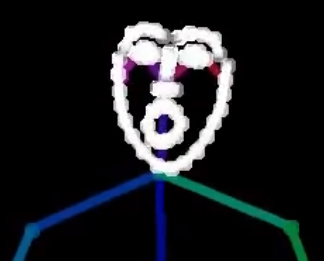}
  \caption{One frame generated by TTS audio when people pause speaking. Mouth shape is distorted.}
  \label{fig:distorted}
\end{minipage}%
\hspace{1cm}%
\begin{minipage}[t]{.45\textwidth}
  \centering
  \includegraphics[width=0.86\linewidth]{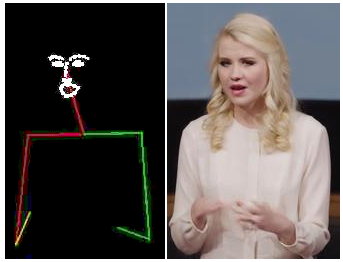}
  \caption{One frame generated by a skeleton model without hands. It is clear that the hand model is necessary to render hand details in the final image.}
  \label{fig:nohand}
\end{minipage}
\end{figure}

\textbf{TTS Noise}.
When we train our LSTM, the audios are extracted from recorded videos, meaning they contain background noise when people are not speaking. However, TTS generated audios have an absolutely clear background when people pause speaking. That difference causes some problems in the output skeleton motions. As can be seen in figure~\ref{fig:distorted}, mouth shape is distorted because our network has never seen this absolutely clear signal in the training. To fix this issue, we add some white noise to the TTS generated audios before feeding to LSTM.

\textbf{Hand Model}.
As mentioned before, it's necessary to have hands in our skeleton model in order to render hand details in the final output of vid2vid. As in figure~\ref{fig:nohand}, we have downloaded a video from Youtube and use it as our training data. Due to its motion blur, we can't fit a correct hand model to the video frames. Thus we trained our vid2vid network without hand skeleton, all the way up to 40 epochs. However, it is still impossible to render clear hand images in the final output. This is also evidence of why the end-to-end approach simply doesn't work. A very detailed spatial guidance is necessary for the GAN network to produce high fidelity rendering. An audio input simply can't provide this spatial guidance. Thus, we eventually give up employing the end-to-end method.

\textbf{Key Pose Insertion}. To justify the effectiveness of our key pose insertion approach, we conducted another user study. In this study, we simply present pairs of synthesized videos with and without inserted key poses. The participants just need to choose which one is more expressive. For all participants, videos with key poses get 80.6\% of the votes compared to 19.4\% for videos without key poses. This demonstrates the necessity of inserting key poses to enrich the expressiveness of speech.

Please check out video results at  \href{https://youtu.be/MUlRtgbGeUs}{https://youtu.be/MUlRtgbGeUs}.

\section{Conclusion}
We proposed a novel framework to generate 
realistic speech videos using the 3D driven approach, while avoiding building 3D mesh models. We built a table of personal key gestures inside the framework to handle the problem of data sparsity and diversity. More importantly, we utilized 3D skeleton constraints to generate body dynamics, which guarantees the pose to be physically plausible.

\medskip
\noindent 
\textbf{Acknowledgment}. This work was supported by  the NSFC grant No. 62001213.

\bibliographystyle{splncs}
\bibliography{egbib}

\end{document}